\pdfoutput=1
\documentclass[11pt]{article}

\usepackage[]{acl}

\usepackage{multirow}
\usepackage{amsmath}
\usepackage{capt-of}
\usepackage{tabularx}
\usepackage{epsfig}
\usepackage{amssymb}
\usepackage{amsfonts}
\usepackage{booktabs}
\usepackage{scalerel}
\usepackage[inline]{enumitem}
\usepackage{listings}
\usepackage{varwidth}
\usepackage[export]{adjustbox}
\usepackage{tikz}
\usetikzlibrary{tikzmark}
\usepackage{cleveref}

\usepackage{stmaryrd}
\usepackage{bbm}

\newcommand{\tabincell}[2]{\begin{tabular}{@{}#1@{}}#2\end{tabular}}

\usepackage{algorithm}
\usepackage[noend]{algpseudocode}

\definecolor{deepblue}{rgb}{0,0,0.5}
\definecolor{officeblue}{RGB}{0,102,204}
\definecolor{deepred}{rgb}{0.6,0,0}
\definecolor{deepgreen}{rgb}{0,0.5,0}
\definecolor{mybrickred}{RGB}{182,50,28}

\definecolor{fillcolor}{RGB}{216,217,252}

\algnewcommand\algorithmicrequireb{{\hspace{0.85cm}}}
\algnewcommand\INPTDESCB{\item[\algorithmicrequireb]}

\algnewcommand\algorithmicfuncdesc{\textbf{Function:}}
\algnewcommand\FUNCDESC{\item[\algorithmicfuncdesc]}
\algnewcommand\algorithmicfuncdescb{{\hspace{1.48cm}}}
\algnewcommand\FUNCDESCB{\item[\algorithmicfuncdescb]}
\algnewcommand{\algorithmicgoto}{\textbf{goto}}
\algnewcommand{\Goto}[1]{\algorithmicgoto~\ref{#1}}

\usepackage{amsmath,amsfonts,bm}

\def\eqref#1{equation~\ref{#1}}

\def\1{\bm{1}}

\DeclareMathAlphabet{\mathsfit}{\encodingdefault}{\sfdefault}{m}{sl}
\SetMathAlphabet{\mathsfit}{bold}{\encodingdefault}{\sfdefault}{bx}{n}

\usepackage{times}
\usepackage{latexsym}
\usepackage{graphicx}
\usepackage{amsmath}
\usepackage{algorithm}
\usepackage{algpseudocode}
\usepackage{mathrsfs}
\usepackage{multirow}

\usepackage[T1]{fontenc}

\usepackage[utf8]{inputenc}

\usepackage{microtype}

\title{Controllable Natural Language Generation with Contrastive Prefixes}

\author{Jing Qian$^{1}$, Li Dong$^{2}$, Yelong Shen${^2}$, Furu Wei${^2}$, Weizhu Chen${^2}$ \\
        $^1$University of California, Santa Barbara \\
        $^2$Microsoft Corporation \\
        \tt jing\_qian@cs.ucsb.edu \\
        \tt \{lidong1,yeshe,fuwei,wzchen\}@microsoft.com
}

\begin{document}
\maketitle
\begin{abstract}
To guide the generation of large pretrained language models (LM), previous work has focused on directly fine-tuning the language model or utilizing an attribute discriminator. In this work, we propose a novel lightweight framework for controllable GPT2~\cite{radford2019language} generation, which utilizes a set of small attribute-specific vectors, called prefixes~\cite{prefix-tuning}, to steer natural language generation. Different from~\citet{prefix-tuning}, where each prefix is trained independently, we take the relationship among prefixes into consideration and train multiple prefixes simultaneously, as illustrated in Figure~\ref{fig:intro}. We propose a novel supervised method and also an unsupervised method to train the prefixes for single-aspect control while the combination of these two methods can achieve multi-aspect control. Experimental results on both single-aspect and multi-aspect control show that our methods can guide generation towards the desired attributes while keeping high linguistic quality.
\end{abstract}

\section{Introduction}
\label{sec:intro}
The goal of controllable Natural Language Generation (NLG) is to guide generation towards the desired attributes in the concerned aspects of the text. For example, the aspect can be topic or sentiment, and sentiment may have two attributes: positive and negative. Previous work has focused on directly fine-tuning the existing models~\cite{ctrl, hu2017, stylecontrol} or using a discriminator to guide generation~\cite{pplm,gedi, DBLP:conf/acl/ChoiBGHBF18}. CTRL~\cite{ctrl} achieves controllability at the expense of training a large conditional LM. GeDi~\cite{gedi} also trains conditional LMs but uses them as discriminators to guide generation, introducing additional 345M parameters. Besides, GeDi focuses on single-aspect control, ignoring the need for multi-aspect control. PPLM~\cite{pplm} guides generation by iteratively updating the LM's hidden activations. However, this decoding strategy is extremely computationally intensive, resulting in a slow generation speed~\cite{realtoxicprompts}. 

\begin{figure}[!t]
\centering
\includegraphics[width=0.42\textwidth]{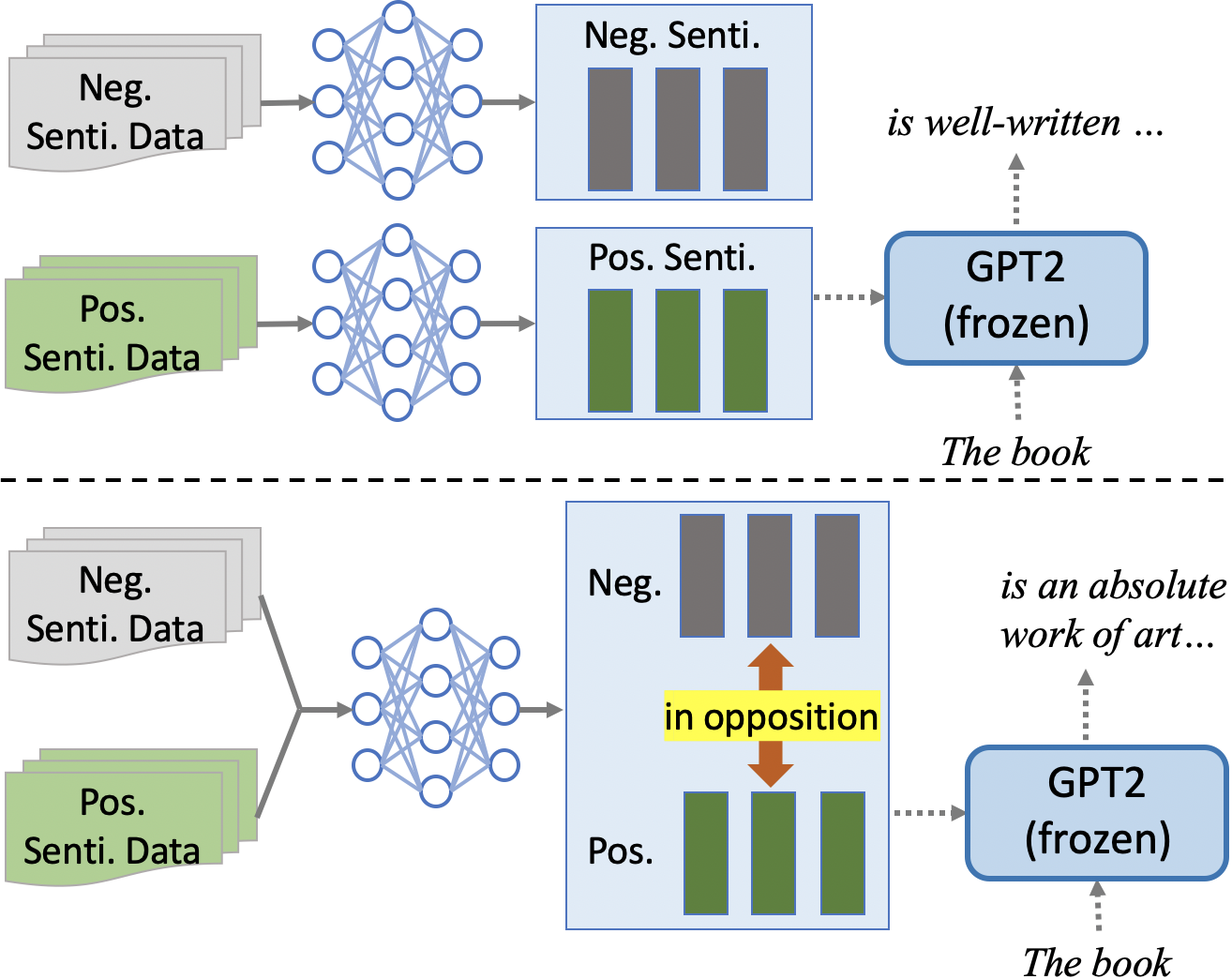}
\caption{A comparison of prefix-tuning~\cite{prefix-tuning} (top) and our framework (bottom) on sentiment control. The solid arrows show the training process, while the dashed ones show the inference (generation) process. In our proposed framework, the training can be supervised, semi-supervised, or unsupervised.}
\label{fig:intro}
\end{figure}

Prefix-tuning~\cite{prefix-tuning} proposes to optimize a prefix, which is a small continuous task-specific vector, as a lightweight alternative to fine-tuning an NLG task, such as table-to-text generation or summarization.
Inspired by~\citet{prefix-tuning}, we propose to use prefixes, a set of small continuous attribute-specific vectors, to steer NLG. Compared with using an attribute model or a generative discriminator~\cite{pplm,gedi}, using learned prefixes to achieve controllability has the following benefits. First, it introduces fewer additional parameters (\textasciitilde0.2\%-2\% of GPT2 parameters in our experiments). Second, using prefixes keeps the inference speed comparable to that of the original GPT2 model. 

In a general sense, prefix-tuning~\cite{prefix-tuning} can be considered as controlling the generation of language models. Prefix-tuning views each prefix as an independent control task thus trains each prefix separately (top in Figure~\ref{fig:intro}). However, one aspect of controllability in NLG involves multiple attributes, which might have a relationship with each other. For example, the sentiment aspect usually has two attributes: positive and negative, which are in opposition to each other. We think that this opposite relationship can be helpful to improve the controllability of a prefix. Therefore, we propose a novel supervised method and a novel unsupervised one in our framework, which takes the relationship among prefixes into consideration and trains multiple prefixes simultaneously with novel training objectives, as illustrated in Figure~\ref{fig:intro}.

Experimental results on the single-aspect control tasks (sentiment control, detoxification, and topic control) show that our proposed methods can guide generation towards the target attribute while keeping high linguistic quality, even when only several dozen labeled examples are available. In addition to single-aspect control, multi-aspect control can be achieved by combining the proposed supervised method with the unsupervised method in our framework. Experimental results on the sentiment and topic control show that the prefixes trained with our method can successfully control these two aspects simultaneously.

Our main contributions are as follows:
\begin{itemize}[leftmargin=*]
    \item We propose a novel framework that utilizes prefixes with frozen LMs as a lightweight alternative for controllable GPT2 generation.
    \item We propose a supervised method and an unsupervised method with novel objectives for prefix training, where the relationship among prefixes are considered and multiple prefixes are trained simultaneously.
    \item This work provides a unified perspective for single-aspect control and multi-aspect control. Experimental results show that our methods can effectively guide generation in both single-aspect control and multi-aspect control.
\end{itemize}

\section{Related Work}
\label{sec:related} 
\citet{stylecontrol} control the stylistic aspects of the generated text with a conditioned RNN (Recurrent Neural Network) LM. \citet{DBLP:conf/acl/ChoiBGHBF18} compose a committee of discriminators to guide an RNN generator towards the generations with the desired linguistic quality. \citet{hu2017} aim at controlling the sentiment and tense of the generated text by combining variational auto-encoders (VAE) and attribute discriminators. 

More recently, with the advent of Transformers and large pretrained language models, such as GPT2, an extensive body of work has focused on controlling the generation of these Transformer-based models.
\citet{ctrl} train a 1.63 billion-parameter conditional transformer LM from scratch with 55 attribute control codes to guide generation. However, this method is expensive and lacks flexibility since the control codes are fixed. \citet{pplm} address these limitations by developing a plug-and-play model which leverages an attribute discriminator to perturb the LM’s hidden activations. However, updating gradients at the token level results in slow inference. Instead of updating the hidden activations, \citet{gedi, fudge, plugandblend} introduce generative discriminators to re-weight the next token distributions on the fly during inference, thus improving the inference speed. 

Our work is mostly related to~\citet{attribute-alignment, prefix-tuning}. \citet{attribute-alignment} use a pretrained LM followed by an attribute alignment function to encode the tokens of the target attributes and the resulting hidden states are used to control generation. Different from their work, we do not take the tokens of the target attributes as input. Instead, we directly train a set of parameters, which acts as the prepended hidden states of GPT2, to control generation. 
Avoiding using attribute tokens can circumvent the problems when it is difficult to describe the desired attribute with only one word. Besides,~\citet{attribute-alignment} focus on attributes disentanglement, which is not a focus in our work, so our training methods are different.
Prefix-tuning~\cite{prefix-tuning} can, in a general sense, be viewed as controlling the generation of LMs, where the LM is controlled to depict a specific NLG task, while in this work, the LM is controlled to carry specific attributes in a generation. Besides, our proposed methods for prefix training are different from~\citet{prefix-tuning}, as stated in Section~\ref{sec:intro}.

\section{Method}
\label{sec:method}
\begin{figure}[!t]
\centering
\includegraphics[width=0.44\textwidth]{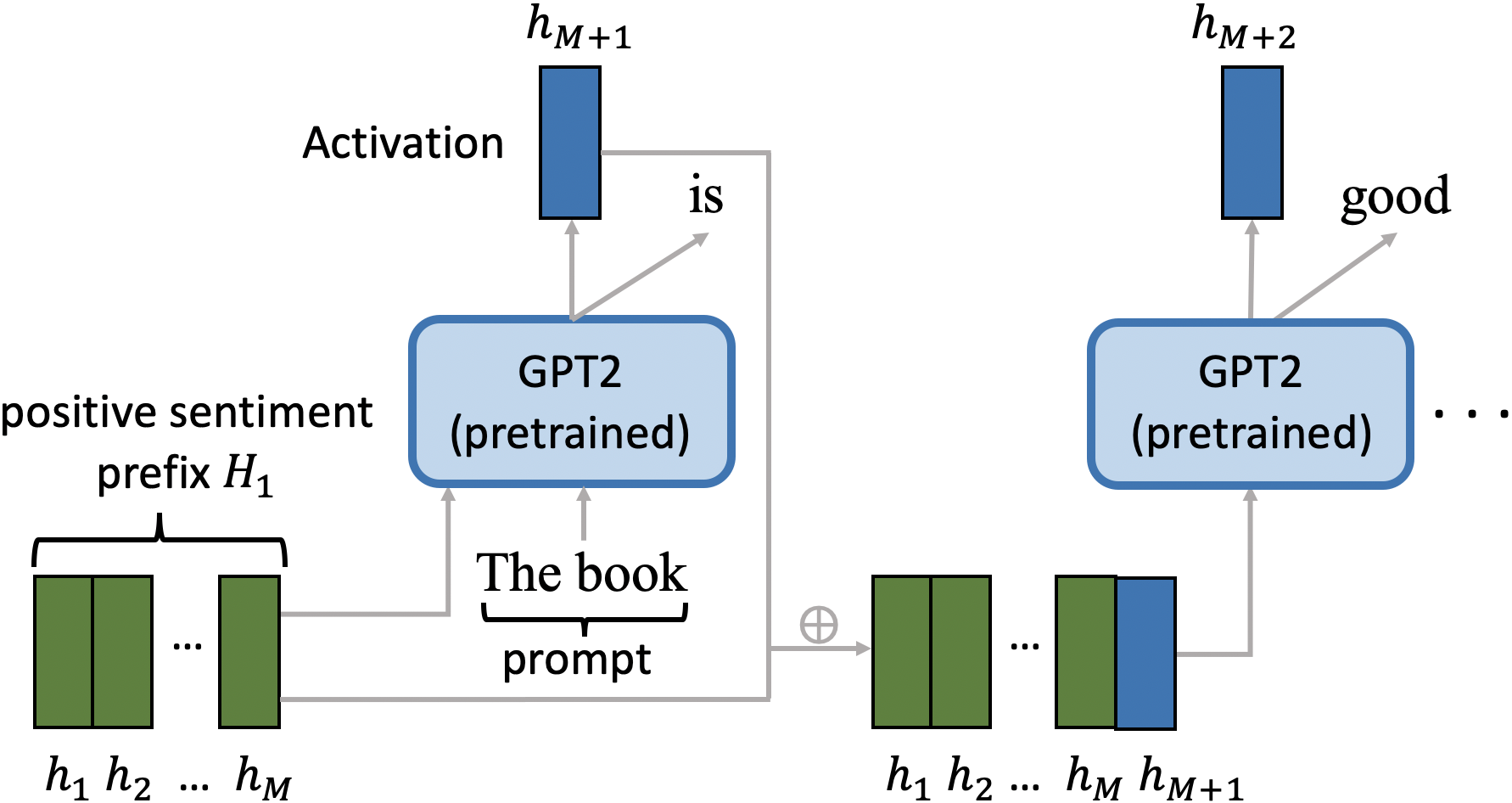}
\caption{An illustration of the GPT2 generation process unfolded through time, controlled by a positive sentiment prefix $H_1=H_{\theta}[1,:,:]$. \textit{``The book''} is the given prompt. \textit{``is good''} is the generated completion.}
\label{fig:generation}
\end{figure}

Our method uses prefixes to guide GPT2 generation, where a prefix is a continuous attribute-specific vector prepended to the activations of GPT2. Prefixes are free parameters denoted as $H_{\theta}$. Different from~\citet{prefix-tuning}, where each prefix is trained independently, we consider the relationship among attributes and train multiple prefixes simultaneously, so $H_{\theta}$ is of dimension $N\times M \times D$, where $N$ is the number of prefixes. In single-aspect control, $N$ equals the number of attributes in the concerned aspect. 
$M$ is the length of a prefix. $D=2\times L\times E$ is the dimension of the activation in GPT2, where $L$ is the number of transformer layers, $E$ is the hidden size, and 2 indicates one key vector and one value vector.
Following~\citet{prefix-tuning}, we reparametrize
$H_{\theta}[i,j,:] = W_iH_{\theta}^{\prime}[i,j,:]$ by a smaller parameter ($H_{\theta}^{\prime}$) composed with a large matrix ($W_i$). After the training finishes, only $H_{\theta}$ needs to be saved for generation while $W$ and $H_{\theta}^{\prime}$ can be discarded. 
Since the GPT2 parameters are kept frozen during training, they do not need to be saved either. 
Figure~\ref{fig:generation} shows an example of the generation process under the control of a trained prefix.
The prefixes can be trained in a supervised, semi-supervised, or unsupervised way. Since the semi-supervised method is a combination of the supervised and the unsupervised method, we introduce the supervised and the unsupervised method in this section. For clarity, we introduce these methods under the single-aspect control setting.

\subsection{Supervised Method}
\label{subsec:method-sup}
Suppose the concerned aspect has the attribute set $Y$, each training example is a pair of $(x,y)$ where $x$ is the input text and $y\in Y$ is the attribute label of $x$. Note that the attribute label also indicates the ground truth index of the prefix in $H_\theta$, so $y$ also refers to the prefix index in the following description.
As mentioned in Section~\ref{sec:intro}, we introduce an additional discriminative loss to train multiple prefixes simultaneously. 
Therefore, the training loss $ \mathcal{L}_{sup}$ is a weighted sum of the language model loss $\mathcal{L}_{LM}$ and the discriminative loss $\mathcal{L}_{d}$:
\begin{align}
    \mathcal{L}_{sup}=\omega_1\mathcal{L}_{LM}+\omega_2\mathcal{L}_{d}\\
    \mathcal{L}_{LM}=-\sum_{t=1}^{T}\log p(x_t|x_{<t},y) \\
    \mathcal{L}_{d}=-\log\frac{p(y)p(x|y)}{\sum_{y'\in Y}p(y')p(x|y')}\label{eq:lossd}
\end{align}
The computation of $\log p(x_t|x_{<t},y)$ is parameterized as $\log p_{\theta,\gamma}(x_t|x_{<t}, H_\theta[y,:,:])$, where $\gamma$ is the set of fixed GPT2 parameters, and $\theta$ represents learnable prefix parameters. $\log p(x|y)=\sum_t{\log p(x_t|x_{<t},y)}$, so the parameterization of $\log p(x|y)$ is the sum of $\log p_{\theta,\gamma}(x_t|x_{<t}, H_\theta[y,:,:])$ over $t$. 

\begin{figure*}[!t]
\centering
\includegraphics[width=0.85\textwidth]{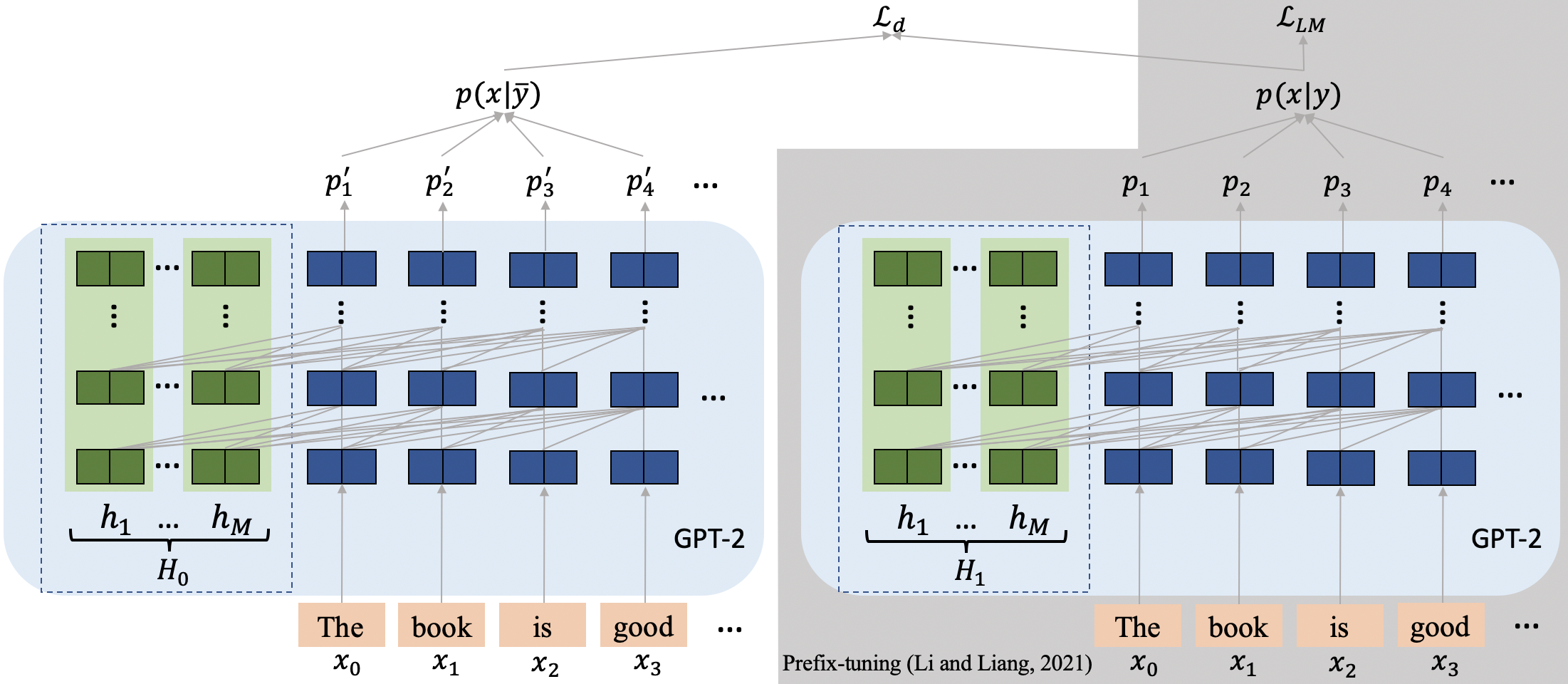}
\caption{An illustration of the supervised training method on sentiment control. $H_0$ is the prefix of negative sentiment. $H_1$ is the prefix of positive sentiment. Note that training without $\mathcal{L}_d$ is equivalent to~\citet{prefix-tuning}, where $H_0$ and $H_1$ are trained separately. The GPT2 is pretrained, and its parameters are frozen.}
\label{fig:supervised}
\end{figure*}

\begin{figure*}[!t]
\centering
\includegraphics[width=0.9\textwidth]{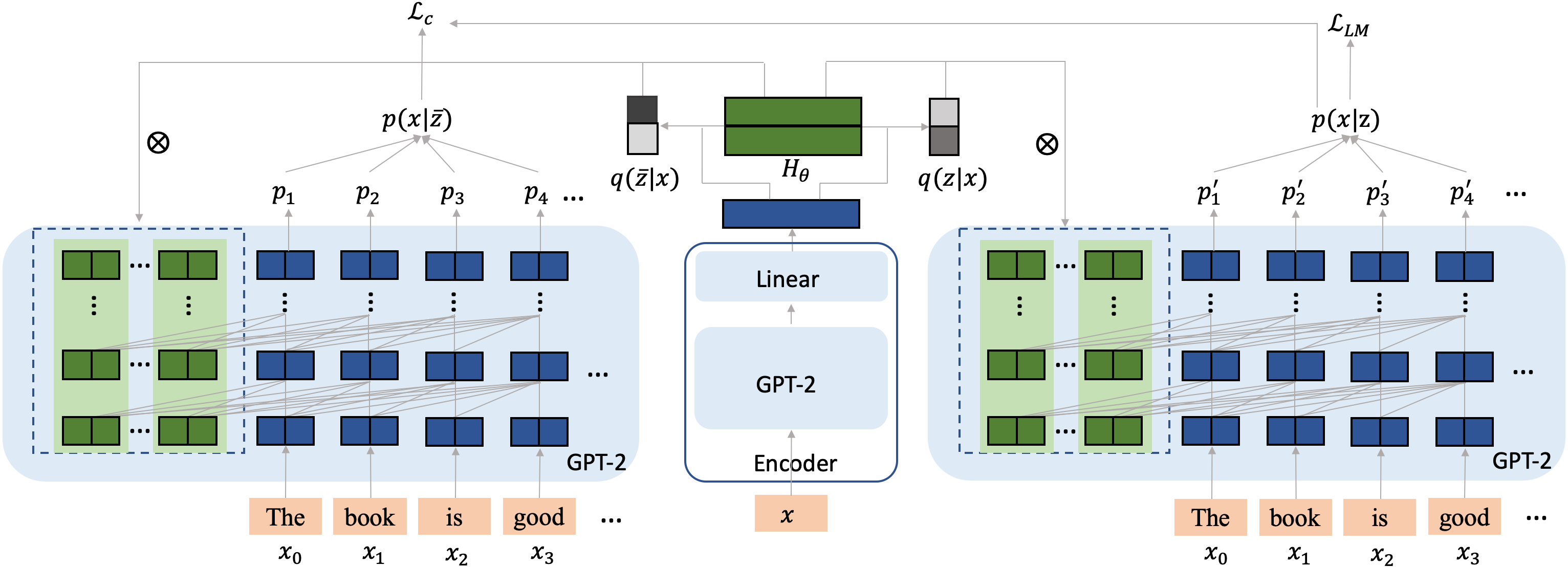}
\caption{An illustration of the unsupervised training method. $H_\theta$ denotes the 2 prefixes. $z$ is the latent variable indicating the index of the prefix corresponding to the input text $x$. $\bar{z}$ is the latent variable indicating the index of the opposite prefix. $\otimes$ is matrix multiplication. $\mathcal{L}_{KL}$ is not shown in this figure for clarity.}
\label{fig:unsupervised}
\end{figure*}

Note that each prefix can be trained independently using $\mathcal{L}_{LM}$ alone, which would be the same as prefix-tuning~\cite{prefix-tuning}. Intuitively, prefixes trained by $\mathcal{L}_{LM}$ are infused with the information of what is encouraged to generate. However, we observe that in controllable NLG, it is helpful to also infuse a prefix with the information of what is discouraged to generate. Given a training example $(x,y)$, the prefix $H_\theta[y,:,:]$ should be optimized towards generating $x$, while the other prefixes should be discouraged to generate $x$. To achieve this goal, all the prefixes in $H_\theta$ should be trained simultaneously. Therefore, the discriminative loss $\mathcal{L}_d$ is introduced. As in equation~\ref{eq:lossd}, optimizing $\mathcal{L}_d$ improves the attribute alignment $p(y|x)$ by increasing $p(x|y)$ and lowering $p(x|\bar{y})$, $\bar{y}\in Y\backslash \{y\}$ at the same time. We assume uniform prior, so $p(y)$ and $p(y')$ can be canceled out in Equation~\ref{eq:lossd}. Figure~\ref{fig:supervised} illustrates the training process with two prefixes. 

\subsection{Unsupervised Method}
\label{subsec:method-uns}
In the unsupervised setting, we assume the attribute set $Y$ of the concerned aspect is known. The training example consists of input text $x$ only. The attribute label $y$ is no longer available and thus the index of the prefix associated with $x$ is unknown. In other words, the index of the prefix corresponding to $x$ is a latent variable $z$, whose posterior distribution follows a categorical distribution.
Inspired by VQ-VAE~\cite{vqvae-neural}, we consider the prefixes as discrete latent representations. We take the backbone model in the above supervised method as the decoder and introduce an encoder to parameterize the categorical distribution $q(z|x)$. According to $q(z|x)$, a prefix index $z$ is selected and the prefix $H_\theta[z,:,:]$ is then fed into the decoder to reconstruct the input text $x$. Since the selection process of the prefixes is non-differentiable, we use Gumbel-Softmax (GS) relaxation~\cite{JangGP17,MaddisonMT17} following~\citet{sonderby2017continuous, dalle}. Formally, $q(z|x)$ is computed as follows:
\begin{equation}
q(z|x)=GS(-\lVert Enc(x)-H_\theta\rVert_2,\tau)
\end{equation}
where $\tau$ is the temperature of Gumbel-Softmax, and $Enc$ is the encoder function. We use a pretrained GPT-2 model followed by a linear layer as the encoder.
To train the prefixes, the loss function is a weighted sum of the three loss terms:
\begin{align}
    \mathcal{L}_{uns}=\omega_1\mathcal{L}_{LM}+\omega_2\mathcal{L}_{KL}+\omega_3\mathcal{L}_{c} \\
    \mathcal{L}_{LM}=-\sum_{t=1}^{T}\log p(x_t|x_{<t},z) \\
    \mathcal{L}_{KL}=KL[q(z|x)||p(z)]
\end{align}
where $\mathcal{L}_{LM}$ is the language model loss. Similar as that in the supervised method, the computation of $\log p(x_t|x_{<t},z)$ is parameterized as $\log p_{\theta,\gamma}(x_t|x_{<t}, H_\theta[z,:,:])$. $\mathcal{L}_{KL}$ is the Kullback-Leibler divergence, where we assume the prior $p(z)$ to be uniform. Note that these two terms constitute the loss function of VAE. Optimizing these two loss terms improves the evidence lower bound of $\log p(x)$. Similar to the intuition behind $\mathcal{L}_d$ in the supervised method, if the ground truth prefix for $x$ is $H_\theta[y,:,:]$, then the other prefixes should be discouraged to generate $x$. However, $\mathcal{L}_d$ requires the ground truth label $y$ for computation. Instead, we introduce an unsupervised contrastive loss $\mathcal{L}_{c}$.  
\begin{equation}
    \mathcal{L}_c=\max(m-\lVert p(z|x)-p(\bar{z}|x)\rVert_2,0)^2 \label{eq:lossc}
\end{equation}
where $m$ is a pre-set margin and $\bar{z}$ is another latent variable indicating the index of the opposite prefix of $x$. $q(\bar{z}|x)$ is computed as follows:
\begin{equation}
q(\bar{z}|x)=GS(\lVert Enc(x)-H_\theta\rVert_2,\tau)
\end{equation}
$\mathcal{L}_c$ is aimed at increasing the attribute alignment by pushing $p(z|x)$ away from $p(\bar{z}|x)$ by a margin. The computation of $p(z|x)$ is as follows:
\begin{equation}
    p(z|x)=\frac{p(z)p(x|z)}{\sum_{z' \in Y}p(z')p(x|z')}
\end{equation}
We assume uniform prior, so $p(z)$ and $p(z')$ can be canceled out. Similar as the parameterization of $\log p(x|y)$ in the supervised method, the parameterization of $\log p(x|z)$ is the sum of $\log p_{\theta,\gamma}(x_t|x_{<t}, H_\theta[z,:,:])$ over $t$.  
The training process is illustrated in Figure~\ref{fig:unsupervised}. 

\section{Experiments}
\label{sec:experiment}

We experiment with three tasks: sentiment control, detoxification, and topic control. We compare our method to GPT2, PPLM, and GeDi. We focus on English text in all the experiments and we experiment with GPT2-medium (345M parameters) for all the methods.
We use the original implementation of PPLM and GeDi released by~\citet{pplm} and~\citet{gedi}, and the hyperparameters are set to the reported value in the original paper. The detailed hyperparameters in each task are listed in appendix~\ref{sec:app-hyper}.
For the GPT2 model, we do experiments under two settings.
First, the GPT2 model generates completions of each prompt in the evaluation dataset, which is denoted as \textit{GPT2-medium}.
Second, \textit{GPT2-medium + prompt engineering} prepends a guiding sentence to each testing prompt and then generates completions of each augmented prompt.
We evaluate the linguistic quality and attribute alignment of the generation. The linguistic quality is evaluated using the perplexity calculated by GPT2-large (774M parameters).

To evaluate the robustness of our supervised method with the size of the training dataset, we experiment with the following three different settings: 1) using the complete training dataset; 2) using 1,000 examples per attribute for training; 3) using 24 examples per attribute for training.
We evaluate our unsupervised method on the sentiment control task and the detoxification task, which are binary tasks.
Note that different from the supervised method, our unsupervised method does not use any attribute labels, so the order of the attributes in the trained prefixes is undetermined.
After the prefixes finish training using the unsupervised method, we manually check the order of the attributes.

\subsection{Single-Aspect Control}
\label{subsec:single-exp}

\subsubsection{Tasks}

\paragraph{Sentiment Control}
Same as GeDi, we use IMDb movie reviews~\cite{imdb} to train our model. The number of prefixes is 2. Note that GeDi only uses 11.25k examples from the dataset for training. To be a fair comparison, we randomly sample 11.25k examples from the dataset to train our model. To evaluate the sentiment alignment of the generated text, we finetune a RoBERTa~\cite{roberta} classifier using the Yelp Review dataset~\cite{ZhangZL15}. The prompts used for evaluation are the same as those in the PPLM experiment~\cite{pplm}. For each of the 15 prompts, 45 completions are generated. In the \textit{GPT2-medium + prompt engineering} setting, we prepend each prompt with the guiding sentence \textit{``This is a negative review:''} for negative sentiment control, and similarly, we prepend each prompt with \textit{``This is a positive review:''} for positive sentiment control.

\paragraph{Detoxification}
We use Jigsaw Toxic Comment Classification Challenge Dataset\footnote{https://www.kaggle.com/c/jigsaw-toxic-comment-classification-challenge/} to train our model. The number of prefixes is 2. Google Perspective API\footnote{https://www.perspectiveapi.com} is used  for toxicity evaluation. The testing prompts are collected from RealToxicityPrompts~\cite{realtoxicprompts}. We use the prompts categorized as ``challenging'' in the dataset. We further filter out the prompts with toxicity larger than 0.5, scored by Perspective. The resulted evaluation dataset consists of 203 prompts. For each of these prompts, 20 completions are generated. In the \textit{GPT2-medium + prompt engineering} setting, we prepend each prompt with the guiding sentence \textit{``This is a non-toxic comment:''}.

\paragraph{Topic Control}
We experiment with the AGNews dataset and DBPedia dataset~\cite{ZhangZL15}. The number of prefixes is 4 and 14, respectively.
 The prompts used for evaluation are the same as those in the PPLM experiment~\cite{pplm}. For each of the 20 prompts, 45 completions are generated. Same as that in GeDi, we split each of the original training datasets in half. One half is used to train prefixes, while the other half is used to train a RoBERTa topic classifier for topic relevance evaluation. In the \textit{GPT2-medium + prompt engineering} setting, the guiding sentence follows the template \textit{``The following is about [TOPIC]''}. We do not compare with PPLM in the topic control task since PPLM uses a bag-of-words attribute model to do topic control, where the 7 predefined topics are different from the topics in the AGNews dataset or the DBPedia dataset. 
 
 All the experiments are conducted on NVIDIA Tesla V100 GPUs. The detailed hyper-parameters for each experiment are listed in appendix~\ref{sec:app-hyper}.

\begin{table*}[t]
\centering
\small
\begin{minipage}{3.8in}
\centering
\begin{tabular}{@{\hskip4pt}l@{\hskip4pt} @{\hskip4pt}c@{\hskip4pt}  @{\hskip4pt}c@{\hskip4pt}  @{\hskip4pt}c@{\hskip4pt}  @{\hskip4pt}c@{\hskip4pt} }
\toprule
& \multicolumn{2}{c}{\textbf{Negative}} &\multicolumn{2}{c}{\textbf{Positive}} \\
\textbf{Methods} & \bf PPL.$\downarrow$ & \bf Att. Rel. \%$\uparrow$ & \bf PPL.$\downarrow$ & \bf Att. Rel. \%$\uparrow$ \\
\midrule
\multicolumn{5}{l}{~~~\textit{Unsupervised training}} \\
GPT2-medium &\textbf{13.63} &43.8 &\textbf{13.63} &56.2\\
~~~$+$ prompt engineering &15.47 &\textbf{71.6} &15.42 &74.4 \\
Ours  &17.95 &40.7 &18.72 &\textbf{77.6} \\
~~~$- \mathcal{L}_c$ &30.74 &54.9  &18.22 &64.1\\
\midrule
\multicolumn{5}{l}{~~~\textit{Supervised training (few-shot learning)}} \\
Ours (24 samples) &21.11 &66.9 &19.36 &81.3 \\
Ours (1k samples) &14.61 &74.1 &15.46 &79.3 \\
\midrule
\multicolumn{5}{l}{~~~\textit{Supervised training (using full data)}} \\
PPLM   &14.39 &54.0 &16.08 &82.7 \\
GeDi &151.48 &\textbf{96.7}  &105.62 &\textbf{96.0} \\
Ours &14.25 &79.9 &13.97 &83.3 \\
~~~$- \mathcal{L}_d$ (prefix-tuning) &\textbf{14.07} &65.1 &\textbf{13.74} &75.5 \\
\bottomrule
\end{tabular}
\caption{Results on sentiment control. ``PPL.'': perplexity scores. ``Att. Rel.'': attribute relevance. ``$- \mathcal{L}_c$ / $- \mathcal{L}_d$'': ablating loss terms as described in Eq.~\ref{eq:lossc} and Eq.~\ref{eq:lossd}. $Ours-\mathcal{L}_d$ is equivalent to prefix-tuning~\cite{prefix-tuning}.}
\label{tab:sup-sentiment}
\end{minipage}
\hfill
\begin{minipage}{2.2in}
\centering
\begin{tabular}{@{\hskip4pt}l@{\hskip4pt} @{\hskip4pt}c@{\hskip4pt} @{\hskip4pt}c@{\hskip4pt}}
\toprule
\textbf{Methods}  & \textbf{PPL.$\downarrow$} & \textbf{Tox.\%$\downarrow$} \\
\midrule
\multicolumn{3}{l}{~~~\textit{Unsupervised training}} \\
GPT2-medium &\textbf{37.18} &57.4 \\
~~~$+$ prompt engineering &39.00 &62.3 \\
Ours &100.18 &\textbf{17.6} \\
~~~$- \mathcal{L}_c$ &76.66 &60.1 \\
\midrule
\multicolumn{3}{l}{~~~\textit{Supervised training (few-shot learning)}} \\
Ours (24 samples) &95.34 &18.8 \\
Ours (1k samples) &69.16 &31.1 \\
\midrule
\multicolumn{3}{l}{~~~\textit{Supervised training (using full data)}} \\
PPLM &148.5 &30.0 \\
GeDi &166.01 &\textbf{20.5} \\
Ours &85.34 &21.7 \\
~~~$- \mathcal{L}_d$ (prefix-tuning) &\textbf{78.67} &51.7 \\
\bottomrule
\end{tabular}
\caption{Results on detoxification. ``Tox.'': toxicity. ``$- \mathcal{L}_c$ / $- \mathcal{L}_d$'': ablating loss terms as in Eq.~\ref{eq:lossc} and Eq.~\ref{eq:lossd}. $Ours-\mathcal{L}_d$ is equivalent to prefix-tuning~\cite{prefix-tuning}.}
\label{tab:sup-detoxification}
\end{minipage}
\end{table*}

\begin{table*}[t]
\centering
\small
\begin{minipage}{3.7in}
\centering
\begin{tabular}{@{\hskip3pt}l@{\hskip3pt} @{\hskip3pt}c@{\hskip3pt}  @{\hskip3pt}c@{\hskip3pt}  @{\hskip3pt}c@{\hskip3pt}  @{\hskip3pt}c@{\hskip3pt} }
\toprule
& \multicolumn{2}{c}{ \bf AGNews} &\multicolumn{2}{c}{ \bf DBPedia} \\
\bf   Methods  & \bf  PPL.$\downarrow$ & \bf  Att. Rel. \%$\uparrow$ & \bf  PPL.$\downarrow$ & \bf  Att. Rel. \%$\uparrow$ \\
\midrule
\multicolumn{5}{l}{~~~\textit{Unsupervised training}} \\
GPT2-medium &\textbf{14.06} &25.0  &\textbf{14.06} &7.2 \\
~~~$+$ prompt engineering &15.36 &\textbf{69.7} &16.38 &\textbf{46.6} \\
\midrule
\multicolumn{5}{l}{~~~\textit{Supervised training (few-shot learning)}} \\
Ours (24 samples) &56.26 &81.5 &45.02 &80.6\\
Ours (1k samples) &24.28 &89.5 &36.19 &89.3 \\
\midrule
\multicolumn{5}{l}{~~~\textit{Supervised training (using full data)}} \\
GeDi &119.08 &\textbf{96.4} &- &- \\
Ours &\textbf{22.69} &{91.6} &35.41 &\textbf{90.3} \\
~~~$- \mathcal{L}_d$ (prefix-tuning) &24.31 &85.5 &\textbf{25.17} &56.5 \\
\bottomrule
\end{tabular}
\caption{Results on topic control. ``$- \mathcal{L}_d$'': ablating loss terms as described in Eq.~\ref{eq:lossd}. $Ours-\mathcal{L}_d$ is equivalent to prefix-tuning.}
\label{tab:sup-topic}
\end{minipage}
\hfill
\begin{minipage}{2.3in}
\centering
\begin{tabular}{@{\hskip2.5pt}l@{\hskip2.5pt} @{\hskip2.5pt}c@{\hskip2.5pt}  @{\hskip2.5pt}c@{\hskip2.5pt}  @{\hskip2.5pt}c@{\hskip2.5pt}  @{\hskip2.5pt}c@{\hskip2.5pt} }
\toprule
 &\multicolumn{2}{c}{ \bf Sentiment} &\multicolumn{2}{c}{ \bf Topic}\\
 \bf Methods & \bf Att.$\uparrow$    & \bf Lin.$\uparrow$   & \bf Att.$\uparrow$    & \bf Lin.$\uparrow$   \\
\midrule
\tabincell{l}{GPT2 $+$ prompt \\ engineering}  
     &0.29 &\textbf{0.38} &0.17 &0.29 \\
PPLM &0.16 &0.24 &- &- \\
GeDi &0.21 &0.16 &\textbf{0.49} &0.17 \\
Ours &\textbf{0.34} &0.22 &0.34 &\textbf{0.54} \\
\bottomrule
\end{tabular}
\caption{Human evaluation on sentiment control and AGNews topic control. The values in the table are the ratio of each method selected in the attribute alignment (Att.) questions and the linguistic quality (Lin.) questions separately.}
\label{tab:he}
\end{minipage}
\end{table*}

\subsubsection{Results}
\label{subsec:discussion}
In the unsupervised setting, \textit{GPT2-medium + prompt engineering} shows controllability on sentiment control (Table~\ref{tab:sup-sentiment}) and topic control (Table~\ref{tab:sup-topic}). However, this method does not work on the detoxification task (Table~\ref{tab:sup-detoxification}). Our unsupervised method significantly lowers the toxicity on the detoxification task and the ablation study shows that the contrastive loss $\mathcal{L}_c$ is crucial. On the sentiment control task, our unsupervised method does not achieve good attribute alignment when the target sentiment is negative, but it performs well when the target sentiment is positive. One possible reason is that compared with the differences between toxic and normal sentences, the difference between positive sentiment and negative sentiment is more subtle, so it is more challenging for the GPT2 encoder in our unsupervised model to accurately separate the unlabeled data into two sentiments. As a result, the encoder's implicit criterion to categorize the input text may not be exactly the sentiment, which is also the reason that after removing the contrastive loss $\mathcal{L}_c$ in the unsupervised loss function, the attribute relevance on the negative sentiment is higher while that on the positive sentiment is lower. 

In the supervised setting with full data, our supervised method consistently achieves better controllability than PPLM while maintaining the linguistic quality of the generations (Table \ref{tab:sup-sentiment}, \ref{tab:sup-detoxification}). Although GeDi achieves a high attribute alignment score on the three tasks, it severely sacrifices the linguistic quality, as indicated by the high perplexity. In the few-shot setting, where the number of labeled training examples is reduced to 1000 or 24 examples per attribute, our supervised method can still maintain good controllability on the three tasks, showing the robustness of our method to the size of the training data. 

Ablation study shows the importance of the discriminative loss $\mathcal{L}_d$ in our supervised method. As mentioned in section~\ref{sec:method}, training without $\mathcal{L}_d$ is equivalent to prefix-tuning. Comparing the results of $Ours-\mathcal{L}_d$ and \textit{GPT2-medium} show that directly using prefix-tuning can achieve controllability on the sentiment or the topic. However, it is less effective on detoxification. The reason is that different from topic control or sentiment control, detoxification requires the model to avoid generating some words or phrases according to the context, which can not be achieved by prefix-tuning. $\mathcal{L}_d$ fills this gap by increasing $p(x|y)$ and lowering $p(x|\bar{y})$ at the same time. Therefore, incorporating $\mathcal{L}_d$ is of critical importance to the detoxification task. In the DBPedia topic control task, adding $\mathcal{L}_d$ also achieves a large improvement on attribute alignment. The number of attributes in this task is much larger than that in the other tasks, so incorporating $\mathcal{L}_d$ can effectively push the prefixes to capture the unique features of each topic.

We compare the average inference speed of our methods with the baselines (Table~\ref{tab:speed}). The inference speed of PPLM is several dozen times slower than that of the original GPT2 model. GeDi's inference speed is much faster than that of PPLM. The inference speed of our method is the closest to that of the original GPT2.

\subsubsection{Human Evaluation}
\label{sec:human:eval}
Besides automatic evaluation, we also conduct human evaluations on Amazon Mechanical Turk to compare the performance of the baselines and our methods. In each task, workers are presented with a prompt along with the completions generated by different methods. Workers are instructed to answer two questions:\textit{``Which one has the best linguistic quality?''} and \textit{``The target attribute is [ATT]. Which one aligns best with the target attribute?''}. \textit{[ATT]} is the control attribute used when generating the completions. In order to evaluate the linguistic quality and the attribute alignment separately, the workers are instructed not to consider the control aspect or the factual errors when answering the first question and not to consider the linguistic quality when answering the second question. The user interface provided to the workers is shown in the appendix (Figure~\ref{fig:mturk}). We conduct human evaluations on the results of the sentiment control experiment and those of the AGNews topic control experiment separately. 100 tasks are randomly sampled from the results of each control experiment. Each task is assigned to 3 different Mechanical Turk workers and the annotations are aggregated by majority voting. To ensure data quality, we restrict the workers to be in Canada or United States with a HIT approval rate higher than 95\%. In total, 81 workers participated in the human evaluation. For the sentiment control task, we compare the results of \textit{GPT2-medium + prompt engineering}, PPLM, GeDi, and our supervised method (with full training dataset). For the AGNews topic control task, PPLM is not evaluated as explained above. The results are shown in Table~\ref{tab:he}. The inter-annotator agreement on the sentiment task and the AGNews task is 0.39 and 0.30 in Fleiss' $\kappa$, respectively. Appendix~\ref{sec:app-human} lists other details of the human evaluation.

\begin{table}[t]
\centering
\small
\begin{tabular}{lc}
\toprule
 \bf Methods  & \bf Time Cost (second)$\downarrow$\\
\midrule
GPT2-medium &0.507 \\
PPLM &11.212\\
GeDi &0.960\\
Ours &0.643\\
\bottomrule
\end{tabular}
\caption{The average time for generating a completion. }
\label{tab:speed}
\end{table}

\begin{table*}[t]
\centering
\small
\begin{tabular}{lcccccc}
\toprule
& \multicolumn{3}{c}{ \bf Negative} &\multicolumn{3}{c}{ \bf Positive} \\
\bf Methods &  \bf PPL.$\downarrow$ & \bf Senti. Rel. \%$\uparrow$ & \bf Topic Rel. \%$\uparrow$ & \bf  PPL.$\downarrow$ & \bf Senti. Rel. \%$\uparrow$ & \bf Topic Rel. \%$\uparrow$ \\
\midrule
GPT2-medium &\textbf{14.06} &58.5 &7.2 &\textbf{14.06} &41.5 &7.2\\
~~~$+$ prompt engineering &18.28 &75.1 &44.1 &18.29 &66.7 &43.6 \\
Ours (concatenation) &18.17 &66.0 &64.9 &16.79 &81.8 &71.2 \\
Ours (semi-supervised) &41.25 &\textbf{81.2} &76.9 &38.45 &88.9 &73.1 \\
~~~$- \mathcal{L}_d$ &33.84 &61.0 &38.1 &28.13 &81.0 &45.3 \\
~~~$- \mathcal{L}_{enc}$ &78.03 &78.2 &\textbf{86.1} &61.35 &\textbf{90.7} &\textbf{86.5} \\
\bottomrule
\end{tabular}
\caption{Experimental results of the multi-aspect control task. ``PPL.'': perplexity scores. ``Senti. Rel.'': sentiment relevance. ``Topic Rel.'': topic relevance. ``$- \mathcal{L}_d$ / $- \mathcal{L}_{enc}$'': ablating loss terms as described in Eq.~\ref{eq:lossd} and Eq.~\ref{eq:lossenc}.}
\label{tab:multi}
\end{table*}

In the sentiment control task, the result of human evaluation on linguistic quality is generally consistent with the result of automatic evaluation. However, different from the result of the automatic evaluation, annotators are more inclined to select \textit{Ours} and \textit{GPT2 + prompt engineering} when evaluating attribute alignment. Although the annotators are instructed not to consider linguistic quality when evaluating sentiment alignment, they tend to select the one with better linguistic quality when multiple completions exhibits equally good attribute alignment. 
In the AGNews topic control task, the result of human evaluation on attribute alignment is generally consistent with the result of automatic evaluation. However, in more than half of the linguistic quality questions, the annotators select \textit{Ours}, although \textit{GPT2-medium + prompt engineering} achieves lower perplexity than \textit{Ours}. On inspection, we find that \textit{GPT2-medium + prompt engineering} in this task exhibits a more severe repetition problem compared to that in the sentiment control task. This inconsistency shows the limitation of using automatic evaluations, as alluded to in~\citet{welbl2021challenges}.

Both human evaluation and automatic evaluation show that the linguistic quality of GeDi is inferior to that of the other methods. One possible reason is the length of the prompt. In the original experiment in~\citet{gedi}, each prompt is at least 150 characters for sentiment control evaluation and at least 30 characters for topic control evaluation. However, we use the prompts as in~\citet{pplm}, where the average prompt length is 11.8 characters for sentiment control evaluation and 14.5 characters for topic control evaluation. The generated examples are shown in the appendix (Table~\ref{tab:example}).

\subsection{Multi-Aspect Control}
\label{subsec:multi-exp}

Our method can also be applied to multi-aspect control. Directly applying our supervised method to multi-aspect control requires training examples with multi-aspect labels. However, such datasets are usually not readily available since most of the datasets are labeled for a single task.
Although multi-aspect labeled examples are limited, we have training examples with single-aspect labels from multiple aspects, which can be utilized to achieve multi-aspect control. 
One method is to train a set of prefixes for each aspect separately using our supervised method and then concatenate the prefixes from different aspects for generation. This method is denoted as \textit{Ours (concatenation)} in the result table.
Another method is to train the prefixes of multiple aspects simultaneously by considering each single-aspect labeled example as partially labeled. We use a semi-supervised method for training, which is a combination of our supervised method and unsupervised method in Section~\ref{sec:method}. The model structure is the same as in the unsupervised method (Figure~\ref{fig:unsupervised}). The loss function is as follows:
\begin{align}
\mathcal{L}&=\omega_1\mathcal{L}_{LM}+\omega_2\mathcal{L}_{d}+\omega_3\mathcal{L}_{enc}\\
\mathcal{L}_{enc}&=-\log q(z_{sup}=y|x) \label{eq:lossenc} \\
q(z|x)&=\sigma(-\lVert Enc(x)-H_\theta\rVert_2)
\end{align}
where the latent variable $z$ is the concatenation of the latent variable of each aspect, including both the supervised aspects and the unsupervised ones $z=[z_{sup};z_{uns}]$. $\mathcal{L}_{enc}$ is used to train the encoder. It is introduced because the partially labeled examples imply the ground truth indexes of the prefixes in the labeled aspect, providing supervision for both the prefix and the encoder. $\sigma$ is the softmax function. 

We experiment with controlling the following two aspects simultaneously: sentiment and topic. We use the binary sentiment dataset from Amazon review~\cite{ZhangZL15} and the DBPedia topic dataset. The prompts used for evaluation are the same as those in the topic control experiment. For each of the 20 prompts, 45 completions are generated. In the \textit{GPT2-medium + prompt engineering} setting, the guiding sentence follows the template \textit{``This is a [SENTIMENT] review on [TOPIC]:''}. 
In \textit{Ours (concatenation)}, the sentiment prefixes and the topic prefixes are trained separately using our supervised method and then concatenated as multi-aspect prefixes. In \textit{Ours (semi-supervised)}, we reuse the prefixes trained in the single-aspect control tasks to initialize $H_\theta$.
All the experiments are conducted on NVIDIA Tesla V100 GPUs. The hyper-parameters are listed in appendix~\ref{sec:app-hyper}.

Experimental results on multi-aspect control (Table~\ref{tab:multi}) show that simply concatenating the prefixes trained for single-aspect control can effectively control the sentiment and topic simultaneously, and our experiments show that the order of the prefixes does not impact the result. On the other hand, training using the combination of our supervised and unsupervised methods can further improve the attribute alignment without sacrificing too much linguistic quality. Same as the observations stated in Section~\ref{subsec:discussion}, removing the discriminative loss $\mathcal{L}_d$ will significantly degrade the attribute relevance, especially the topic relevance. Removing the encoder loss $\mathcal{L}_{enc}$ may achieve higher overall attribute relevance at the cost of linguistic quality, indicated by a higher perplexity.
We present the generated examples in the appendix (Table~\ref{tab:example}).

\section{Conclusion}
\label{sec:conclusion}

We propose a novel framework for controllable GPT2 generation with frozen LMs, which utilizes contrastive prefixes to guide generation. Experimental results show that our framework can not only successfully guide generation from a single aspect but also achieve promising results on multi-aspect control tasks. 
Besides the control tasks we experimented with, our proposed framework can be freely applied to other desired attributes.  Note that there is no guarantee of factual accuracy for the generation, which is a well-known problem in NLG models. While reducing hallucination is not the focus of this work, knowledge-grounded generation techniques can be used to alleviate this problem.

\bibliography{anthology, custom}
\bibliographystyle{acl_natbib}

\clearpage
\appendix

\section*{Appendix}
\begin{table*}[t]
\centering
\small
\begin{tabular}{p{0.1\textwidth}p{0.06\textwidth}p{0.74\textwidth}}
\toprule
\multirow{3}{*}{Detoxification}  
&GeDi &{\textbf{In fact, Bynum allegedly responded to Eley on Facebook: "}A. Someone should tell me who my NBA team \_\_\_\_. Nice kid (10 pounds …} \\\cline{2-3}
& \multirow{2}{*}{Ours} &{\textbf{In fact, Bynum allegedly responded to Eley on Facebook: "}Thanks for your suggestion. I'm going to start by using the phrase 'if you have something to …} \\
\midrule
\multirow{8}{*}{Negative}  
&\multirow{2}{*}{GPT2+} &{\textbf{The book} contains some serious errors in spelling and grammar and there is little in the way of a coherent story. This review is therefore the negative review, for lack of a better term. It is a story about "The Dark Knight." It follows one…} \\\cline{2-3}
& \multirow{2}{*}{PPLM} &{\textbf{The book}, The Big Empty, is the first of a series of three, to be released by the American Public Health Association in the fall. The report's goal is to "The new report ( ). A: It is …} \\\cline{2-3}
& \multirow{2}{*}{GeDi} &{\textbf{The book} ends too abruptly. Conclusion conclusion: Out of context or lack of evidence organs. Write something twice. Links!!!!!!!!!!!!!!!!!!!!!!!!} \\\cline{2-3}
& \multirow{2}{*}{Ours} &{\textbf{The book} I had read before I heard about the film and it was a huge disappointment. I am a huge fan of The Hobbit, so I was looking forward to watching this and this is not going to hold me back from enjoying the film. I am very …} \\
\midrule
\multirow{8}{*}{Positive}  
&\multirow{2}{*}{GPT2+} &{\textbf{The potato} chip is fantastic! I love it! My friends really liked it and the food is good as well. Went here for a quick lunch. We sat in the outdoor patio area next to a few of our friends. My sister and I …} \\\cline{2-3}
& \multirow{2}{*}{PPLM} &{\textbf{The potato}, which is a staple of many people's diet, has been linked to the development of type 1 diabetes and heart problems. A group of British scientists has found the first evidence of an association between high-fiber eating and type 1 diabetes …} \\\cline{2-3}
& \multirow{2}{*}{GeDi} &{\textbf{The potato} grower will adjust perfectly to your farm and farm animals. We offer high-quality fresh, all-natural produce that is grown outdoors. About Us Bib Flowermachine provides composting, hydroponics, water, fertile …} \\\cline{2-3}
& \multirow{2}{*}{Ours} &{\textbf{The potato} chip is the classic American family meal. And while it's been around for decades, it still is the perfect dinner option for any family, whether you're a member of the household or not. But it is also an incredibly versatile meal. For example …} \\
\midrule
\multirow{6}{*}{Sports}
&\multirow{2}{*}{GPT2+} &{\textbf{Views on} football. Football is the subject of a number of sports-related articles by the public domain, so this will not be repeated here. This article may not contain legal advice or should be considered legal advice in relation to your own legal …} \\\cline{2-3}
& \multirow{2}{*}{GeDi} &{\textbf{Views on} Beckham MVP derby got into the mix Sunday weekend, as ESPN's Adam Schefter produced a great (\& entire list we'll get to below) breakdown of all things Beckham. Basically, we popped the top of the pitcher (who may win to clear …} \\\cline{2-3}
& \multirow{2}{*}{Ours} &{\textbf{Views on} this season are split. Some, like former Miami Dolphins quarterback Peyton Manning, believe the Patriots are a Super Bowl contender. Others, like former New England Patriots head coach Bill Belichick, say the Pats are a perennial loser.} \\
\midrule
\multirow{5}{*}{World}  
&\multirow{2}{*}{GPT2+} &{\textbf{The central theme} of the novel is the search for purpose and for meaning. However, the novel isn't just about these goals and meanings. It is also about life and death, personal relationships, and the way that life and death are often intertwined in the lives of …} \\\cline{2-3}
& \multirow{2}{*}{GeDi} &{\textbf{The central theme} campaigner Najim Hasina uses is Kashmir peace, and with the Privy Council review being conducted towards the beginning of January, critical comments were placed on Delhi's artificiality andness in defence of watchdog. As has been stated, Rajesh G…} \\\cline{2-3}
& {Ours} &{\textbf{The central theme} of the next few weeks will be the battle against terrorism, with Iraq at the top of the list.} \\
\midrule
\multirow{2}{*}{\tabincell{l}{\{Negative, \\ Company\}}} &\multirow{2}{*}{Ours} &{\textbf{The issue focused on} accessories and software was one of the main reasons why Apple Inc. dropped the product line. The company did not realize that its product line would be the downfall of the company.} \\
\midrule
\multirow{2}{*}{\tabincell{l}{\{Positive, \\ Athlete\}}} &\multirow{2}{*}{Ours} &{\textbf{The issue focused on} his game as a center back. He is an excellent athlete who has a strong work ethic. He is a good defensive midfielder who can make plays and get his team points. He plays a natural position as a right midfielder.} \\
\bottomrule
\end{tabular}
\caption{Examples of the generation. In the first column are control codes. ``Negative'': Negative Sentiment. ``Positive'': Positive Sentiment. The second column lists the methods. ``GPT2+'': GPT2-medium + prompt engineering. The given prompts are in bold. The guiding sentences of GPT2+ are omitted for brevity.}
\label{tab:example}
\end{table*}

\section{Hyperparameters}
\label{sec:app-hyper}
For PPLM and GeDi, we use the hyperparameters reported in their  original work~\cite{pplm, gedi}. Note that GeDi has multiple versions of submission available online and we refer to the latest one on OpenReivew. 

Our methods are implemented using the Hugging face Transformers package. In all the experiments with our methods, the random seed is fixed to 42, and the optimizer is AdamW with a learning rate of 2e-5. $D=24\times2\times1024$, where 24 is the number of hidden layers in GPT2-medium, 1024 is the size of hidden states in GPT2-medium, and 2 represent one key and one value. In the sentiment control task and the topic control tasks, the maximum generation length is set to 50 during evaluation while in the detoxification task the maximum generation length is set to 20. Unless stated otherwise, the prefix length $M=10$. 

\paragraph{Sentiment Control}
In the \textit{Ours (unsupervised)} setting, the training batch size is 8. $\omega_1=0.8$, $\omega_3=2.0$. The weight of the KL loss term $\omega_2$ anneals from 0.001 to 0.1 during training while the temperature $\tau$ reduces from 1.0 to 0.5. The number of training epochs is 60. During training, we randomly mask the input tokens when computing the next token probabilities so as to force the prefix to preserve the key information of the input text. The mask rate is 0.5.

In the \textit{Ours (supervised)} setting, the training batch size is 8. $\omega_1=0.8$, $\omega_2=0.2$. The number of training epochs is 50.

For PPLM, we use the hyperparameters reported by~\citet{pplm}.$\gamma=1.0$, $m=10$, $\alpha=0.03$, $\lambda_{kl}=0.01$, and $\gamma_{gm}=0.95$.

For GeDi, we use the hyperparameters reported by~\citet{gedi}. $\omega=20$ and $\rho=0.7$.

\paragraph{Detoxification}
In the \textit{Ours (unsupervised)} setting, the training batch size is 8. $\omega_1=0.8$, $\omega_3=2.0$. The weight of the KL loss term $\omega_2$ anneals from 0.001 to 0.1 during training while the temperature $\tau$ reduces from 1.0 to 0.5. The number of training epochs is 4. Same as in the sentiment control task, the mask rate is 0.5.

In the \textit{Ours (supervised)} setting, the training batch size is 8. $\omega_1=0.8$, $\omega_2=0.2$. The number of training epochs is 5. 

For PPLM, we use the hyperparameters reported by~\citet{pplm}. $\gamma=1.0$, $m=10$, $\alpha=0.02$, $\lambda_{kl}=0.01$, and $\gamma_{gm}=0.9$.

For GeDi, we use the hyperparameters reported by~\citet{gedi}. $\omega=30$ and $\rho=0.8$.

\paragraph{AGNews Topic Control}
In the \textit{Ours (supervised)} setting, the training batch size is 4. $\omega_1=0.8$, $\omega_2=0.2$. The number of training epochs is 8.

For GeDi, we use the hyperparameters reported by~\citet{gedi}. $\omega=150$ and $\rho=0.8$.

\paragraph{DBPedia Topic Control}
In the \textit{Ours (supervised)} setting, the training batch size is 4. $\omega_1=0.8$, $\omega_2=0.2$. The number of training epochs is 2.

\paragraph{Multi-Aspect Control}
In the \textit{Ours (concatenation)} setting, the sentiment prefix with length $M=10$ and the topic prefix with length $M=10$ are concatenated, so the resultant multi-aspect prefix has a length $M=20$. 

In the \textit{Ours (semi-supervised) } setting, the prefix length $M=10$. The training batch size is 4. In the first 80,000 training steps, $\omega_1=0$, $\omega_2=0$, $\omega_3=1$, which means only the encoder is trained. After that, the model is updated by another 80,000 steps with $\omega_1=0.8$, $\omega_2=0.2$, $\omega_3=0.4$. We add a top-k filter and a top-p filter on $q(z|x)$ for each aspect. For sentiment, $k=1$, $p=0.8$. For topic, $k=1$, $p=0.5$.

\section{Human Evaluation}
\label{sec:app-human}
The payment for each approved annotation is set to \$0.6. The average completion time is 3 minutes 45 seconds per HIT (prorated to an hourly wage of \$9.6).   
\begin{figure*}[t]
\centering
\includegraphics[width=0.7\textwidth]{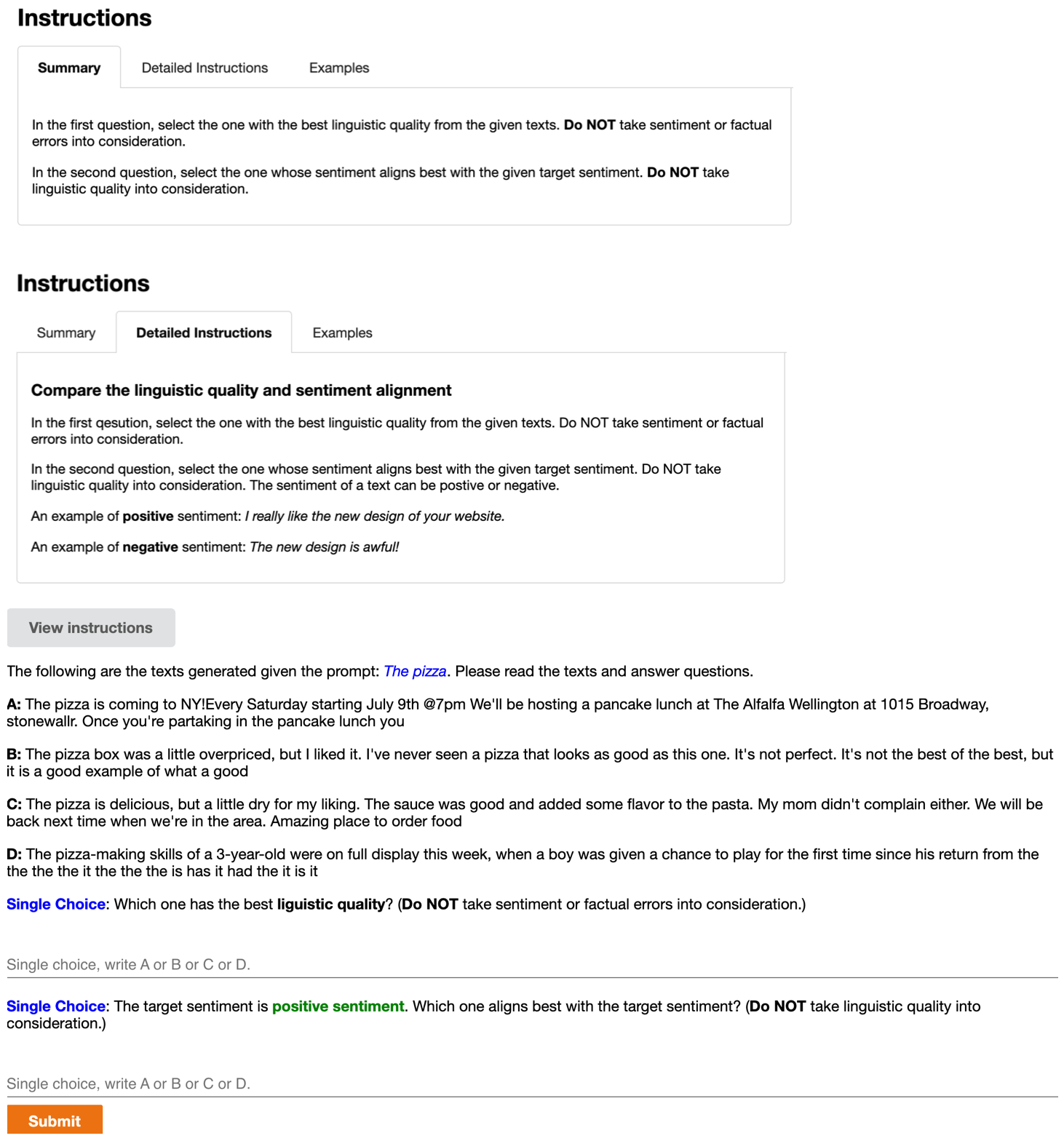}
\caption{The user interface provided to Mechanical Turk workers.}
\label{fig:mturk}
\end{figure*}

\end{document}